\pdfoutput=1
\documentclass[11pt]{article}

\usepackage[]{acl}

\usepackage{times}
\usepackage{latexsym}
\usepackage{amsmath}
\usepackage{mathtools}

\usepackage{tikz-dependency}

\usepackage{multirow}

\usepackage[T1]{fontenc}
\usepackage[utf8]{inputenc}

\usepackage{microtype}
\usepackage{amsfonts}

\title{Using dependency parsing for few-shot learning in distributional semantics}

\author{Ștefania Preda \\
  University College London \\ United Kingdom\\
  \texttt{stefipredacs@gmail.com} \\\And
  Guy Emerson \\
  University of Cambridge \\ United Kingdom \\
  \texttt{gete2@cam.ac.uk} \\}

\begin{document}
\maketitle
\begin{abstract}

In this work, we explore the novel idea of employing dependency parsing information in the context of few-shot learning, the task of learning the meaning of a rare word based on a limited amount of context sentences. Firstly, we use dependency-based word embedding models as background spaces for few-shot learning. Secondly, we introduce two few-shot learning methods which enhance the additive baseline model by using dependencies. 

\end{abstract}

\section{Introduction}
\label{sec:intro}

Distributional semantics models create word embeddings based on the assumption that the
meaning of a word is defined by the contexts it is used in
(for an overview, see: \citealp{sahlgren2008distributional,lenci2018distributional,boleda2020distributional,emerson-2020-goals}).
A fundamental challenge for these approaches is the difficulty of producing high-quality embeddings for rare words, since the models often require vast amounts of training examples \citep{adams-etal-2017-cross,van-hautte-etal-2019-bad}.
To address this problem, various few-shot learning methods have been previously introduced. The goal
of a few-shot learning technique is to learn an embedding that captures the meaning of a
word, given only a few context sentences. The rare word's vector has to be placed in an existing \textit{background} space of embeddings.

Few-shot learning in distributional semantics is a relatively underexplored area,
with important practical applications. Having good representations of rare words is highly
desirable in applications aiming to understand dialects or regionalisms, as well as specific technical language.

\begin{figure*}
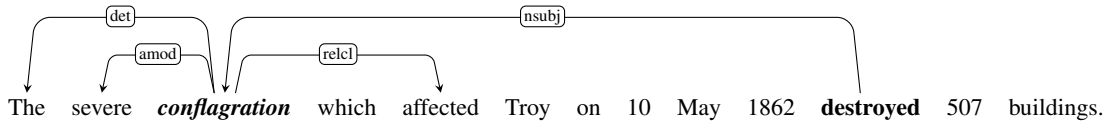

    \centering \small
\begin{dependency}[theme = default]
   \begin{deptext}[column sep=0.2cm]
      The  \& severe \& \textbf{\textit{conflagration}} \& which \& affected \& Troy \& on \& 10 \& May \& 1862 \& \textbf{destroyed} \& 507 \& buildings. \\
   \end{deptext}
   \depedge[edge height=5mm]{3}{2}{amod}
   \depedge[edge height=10mm]{3}{1}{det}
   \depedge[edge height=5mm]{3}{5}{relcl}
   \depedge[edge height=10mm]{11}{3}{nsubj}

\end{dependency}

    \caption{A dependency parse, illustrating that context words connected by a dependency can be important for inferring meaning in a few-shot setting, such as ``destroyed'' for the rare word ``conflagration''.}
    \label{fig:parse}
\end{figure*}

In this work, we explore the idea of incorporating information from the dependency parse of sentences in the context of few shot-learning. An intuition why this might be useful is provided in Figure \ref{fig:parse}. In the given sentence, the most relevant word for inferring the meaning of the target rare word ``conflagration" is ``destroyed". Even if this word is located far from the target, it is directly connected to it through a nominal subject dependency. Moreover, the fact that the target word is used in a certain dependency structure might reveal important characteristics related to its meaning. Since in the case of few-shot learning the data is limited, using dependency parsing information is a resource with great potential to boost existing models.

As a first effort in this direction, this work provides three contributons. Firstly, we explore the effect of using dependency-based word embeddings as background spaces. Secondly, we introduce new few-shot learning methods leveraging the dependency parsing information. Lastly, we update a previous dependency-based background model to make it more suitable for few-shot learning.

\section{Background: dependency-based word embeddings}
\label{sec:dependency}

The widely-used Skip-Gram model introduced by \citet{mikolov} takes the contexts of a word to be those words surrounding it in a pre-defined window size. The model learns the embeddings in an unsupervised manner, using a feed-forward neural network trained on
large amounts of sentences.

\citet{levy-goldberg-2014-dependency} proposed a different way to construct the contexts of a target
word in the training process of the Skip-Gram model. Instead of taking the words from a
pre-defined window, one takes the words that are connected to the target word by a syntactic
dependency. The contexts were defined as the concatenation of the connected word and the
label of the dependency. This allowed the model to differentiate between same words used
in different syntactic roles.

The dependency-based word embeddings were found to be better at capturing similarity, while the window-based models capture relatedness. For example, a dependency-based model would produce close embeddings for ``Rome” and ``Florence”, which are \textit{syntactically similar}
since they can be used in the same grammatical contexts, while a window-based model is likely to place closely the embedding of highly related terms such as ``Rome" and ``ancient", even if they cannot be used interchangeably since they are different parts of speech.

\citeauthor{levy-goldberg-2014-dependency}’s model successfully captured syntactic similarity, but failed to express
how different dependency types affect relations between words. Moreover, it introduced sparsity issues.
\citet{czarnowska-etal-2019-words}
developed the Dependency Matrix model to address these shortcomings. Instead of incorporating the dependency labels in the context vocabulary, each dependency type $d$ is associated with a matrix $T_d$, which
acts as a meaning representation of the link between the target and the context words. The matrices $T_d$, as well as the vectors holding the target vectors $e$ and context vectors $o$, are learned during training. Let $\mathcal{D}$ be the set of training examples given by tuples of target word $t$, context word $c$ and dependency type $d$. For each tuple, we generate a set $\mathcal{D'}$ of negative samples $(t, c', d)$ by drawing context words $c'$ from a noise distribution and maintaining the same target word $t$ and dependency type $d$. The learning goal is to maximise the function in (\ref{eq:1}), where $\sigma$ is the sigmoid function and $e_t$ and $o_c$ are the vectors of the target and context word.%, respectively.
\begin{align}
    \sum_{\mathclap{(t, c, d) \in \mathcal{D}}} \Big ( \log\sigma(u^{t,c,d}) &+ \sum_{\mathclap{(t, c', d) \in \mathcal{D'}}} \log\sigma( - u^{t,c',d}) \Big ) \label{eq:1}\\
    u^{t, c, d} &= e_t \cdot T_{d} \cdot o_c\label{eq:2}
\end{align}

\section{Background: few-shot learning}
\label{sec:few-shot}

As a straight-forward yet successful baseline, the vector
of the rare word is estimated by the sum of the vectors of the words in contexts, as proposed by \citet{lazaridou} and \citet{herbelot-baroni-2017-high}. The latter noticed that not including the stop-words greatly improves the performance on the evaluation tasks. To optimise the performance of the additive model, \citet{van-hautte-etal-2019-bad} proposed weighting the context words according to distance and frequency, as well as subtracting a ``negative sampling'' vector. These modifications take hyperparameters that are important for Skip-Gram's strong performance, such as number of negative samples $k$ and window size $n$ \citep{levy-etal-2015-improving},
and apply them to the few-shot setting. For each word $w$ in the vocabulary $V$, with frequency $f(w)$ and distance $m$ from the target rare word $t$, and for a frequency threshold $\tau$, we calculate the subsampling weight $s(w)$, the window weight $r(w)$ and negative sampling coefficient $n(w)$. 
\vspace*{-2mm}
\begin{align}
    s(w) &= \min\left(1,\sqrt{\frac{\tau}{f(w)}} \label{eq:sub}\right)\\
    r(w) &= \max \Big (0, \frac{n-m+1}{n} \Big) \label{eq:window}\\
    n(w)&= \frac{f(w)^{0.75}}{\sum_{w \in V} f(w)^{0.75}}\label{eq:neg}
\end{align}

Assume $\mathcal{C}$ is the collection of non-stop context words for the given target rare word $t$ and $v_c$ is the vector in the background space for each $c \in \mathcal{C}$. The vector of the target rare word $t$ will is: 
\vspace*{-1mm}
\begin{align}
    v_t &= \sum_{c\in C} v_c^{\text{add}} \textrm{~~~where }\\
    v_c^{\text{add}} &= s(c)  r(c) \Big (v_c - k \sum_{w \in V} n(w) v_{w} \Big ) \label{eq:7}
\end{align}

More involved models have been proposed for the task of few-shot learning. \citet{khodak-etal-2018-la} introduced A La Carte, which applies a linear transformation to
the sum of the context words obtained by the additive model. The weights of the linear transformation are optimised based on the co-occurrence matrix of the corpus. \citet{van-hautte-etal-2019-bad} takes this approach further in the Neural A La Carte model, by using a neural network with a hidden layer to produce a non-linear transformation matrix, which adds flexibility.

The meaning of a rare word can often be deduced from the word form itself. This information has been leveraged in few-shot learning models. For example, the Form-Context Model \citep{schick2019learning} is a hybrid method  which retrieves the weighted sum between the surface form
embedding of the rare word, obtained using FastText \citep{fasttext} and the context-based embedding, produced using the A La Carte model.

In this paper, we focus on additive methods, which do not require additional training on few-shot learning examples. This keeps the inference fast and in line with the \textit{true few-shot learning} setting proposed by \citet{perez2021true}.
%Therefore, we only compare our novel models against the additive model.   % GE: I think this sounds too negative.

% TODO equations?

\section{Dependency-based FSL methods}
\label{sec:dep-fsl}

Dependency relations proved to be an informative tool in the context of creating distributional semantics models. Based on this success, we introduce two dependency-based few-shot learning methods which build on top of the Additive model. In this section, we assume we have already trained a background space of embeddings $v_i$ for each word $i$. In our setup, we chose to consider only the target embeddings learnt by the aforementioned background models, i.e. $v_i = e_i$. Alternatively, one could use the concatenation of the target and context embeddings. 

\paragraph{Dependency Additive Model}
The starting point of our methods is the assumption that the closer a word is to the target word in the dependency graph, the more relevant it is for inferring the target's meaning, as seen in Figure \ref{fig:parse}.

Our method assigns weights for each word in the sentence by considering the distances from the rare word in the dependency parse. For each context word $c$, let $d_c$ be the number of dependency links from the target rare word $t$ to $c$ in the parse. Note that we consider links in both directions. The inferred vector $v_t$ of the rare word is the weighted sum of the vectors of context words, where the weight $w_c$ of each context word $c$ is given in (\ref{eq:4}). The weight is chosen so that it is inversely proportional to the distance from the target, and we add 1 in order to avoid discarding context words which are far from the target in the dependency tree.

\begin{align}
    v_t &= \sum_{c\in C} w_c v_c^{\text{add}} &\textrm{where } w_c &= 1 + \frac{1}{d_c} \label{eq:4}
\end{align}

Initially, we experimented with simply applying the coefficients $w_c$ on the vectors of the context words $v_c$.~However, a better performance was achieved when we incorporated the the weighting steps in (\ref{eq:7}), so we used $v_c^{\text{add}}$ instead of $v_c$.

\begin{table*}[h!]
\centering
\addtolength{\tabcolsep}{-1pt}
\begin{tabular}{|l|l|c|c|c|c|c|c|c|c|c|c|}
\hline
\multirow{2}{*}{\textbf{Backgr. Model}} & \multirow{2}{*}{\textbf{FSL Model}} & \multicolumn{2}{c|}{\textbf{DN}} & \multicolumn{3}{c|}{\textbf{Chimera}} & \multicolumn{5}{c|}{\textbf{CRW}} \\ \cline{3-12} 
& & \textbf{MRR} & \textbf{MR} & \textbf{L2} & \textbf{L3} & \textbf{L6} & \textbf{1} & \textbf{2} & \textbf{4} & \textbf{8} & \textbf{16} \\ \hline
\multirow{2}{*}{\shortstack[l]{Skip-Gram}} & Additive & 0.010 & 5312 & 0.12 & 0.19 & 0.20 & 0.11 & 0.12 & 0.13 & 0.15 & 0.15 \\ \cline{2-12}
& Dep.~Additive & 0.021 & 4007 & 0.13 & 0.20 & 0.21 & 0.12 & 0.13 & 0.14 & 0.15 & 0.16 \\ \hline
\multirow{2}{*}{\shortstack[l]{\strut Dependency\\ Skip-Gram}} & Additive & 0.023 & 4671 & 0.14 & 0.21 & 0.21 & 0.11 & 0.14 & 0.15 & 0.16 & 0.17 \\ \cline{2-12}
& Dep.~Additive & 0.027 & 3785 & \textbf{0.16} & 0.21 & 0.23 & 0.12 & 0.15 & 0.16 & 0.17 & 0.18 \\ \hline
\multirow{3}{*}{\shortstack[l]{Dependency \\ Matrix}} & Additive & 0.017 & 3367 & 0.13 & 0.23 & 0.25 & 0.15 & 0.17 & 0.20 & 0.22 & 0.22 \\ \cline{2-12} 
& Dep.~Additive & \textbf{0.034} & \textbf{3140} & 0.14 & \textbf{0.26} & 0.29 & \textbf{0.18} & \textbf{0.20} & \textbf{0.22} & \textbf{0.24} & \textbf{0.25} \\ \cline{2-12} 
& DM Additive & 0.019 & 3163 & 0.15 & 0.24 & \textbf{0.31} & 0.16 & 0.20 & 0.20 & 0.21 & 0.22 \\ \hline
\end{tabular}
\caption{\label{tab:results}
Results for different combinations of background and few-shot learning model, on three evaluation datasets. The best result for each column is marked in bold.
Higher is better for all columns except MR.
}
\end{table*}

\paragraph{Dependency Matrix Additive Model}
The Dependency Additive model above does not take into account the type of dependency on each edge in the graph, which, as we have seen, plays an important role in capturing the meaning of the words in relation to each other. We thus devised a strategy to make use of this information. 

\citeauthor{czarnowska-etal-2019-words} proposed the idea of using the learnt dependency matrices of the Dependency Matrix model for the task of semantic composition, by multiplying word embdeddings with matrices over chains of dependencies. We apply the same idea in the context of few-shot learning. More precisely, instead of giving a weight for each vector of a context word, we multiply it with corresponding dependency matrices on the chain of dependencies from the target to the context. To be able to do this based on the original Dependency Matrix model, we would have to take into account that when we advance in the dependency parse, we have to switch between using the context vector (retrieved from $o$) and target vector (retrieved from $e$). 

To simplify this process, we modified the Dependency Matrix model to use only one embedding per word, instead of separate context and target embeddings.\footnote{%
    This cannot be applied to Skip-Gram
    without causing every word to predict itself as a context.
    To allow Skip-Gram to use only one vector per word,
    \citet{zobnin-elistratova-2019-learning}
    propose using an indefinite inner product,
    which corresponds to $T$ in~(\ref{eq:5})
    being a diagonal matrix of $1$s and~$-1$s.
    In a similar vein, \citet{bertolini-etal-2021-representing}
    propose a more radical simplification of the Dependency Matrix model,
    which uses matrices that are non-zero only on the diagonal and off-diagonal. 
} This also reduces the training time of the model. More precisely, we have the same training loss as in~(\ref{eq:1}), but (\ref{eq:2})~is replaced by:
\begin{align}
    u^{t, c, d} &= v_t \cdot T_{d} \cdot v_c\label{eq:5}
\end{align}

Having trained this model, we then make use of the matrices $T_d$, optimised for each dependency type $d$. For the target rare word $t$ and each non-stop context word $c$, Let $D(t, c)$ be the path of dependency types from $t$ to $c$. The vector of the target rare word is calculated as:
\begin{align}
v_t &= \sum_{c\in C} v_c' &
\textrm{where } v_c' &= \left(\prod_{d\in D(t,c)}\!\!\!\!\!T_d \right) v_c^{\text{add}}
\label{eqn:sum}
\end{align}

\section{Experiments}
\label{sec:experiments}

In our setup, we considered three background models: window-based Skip-Gram, dependency-based Skip-Gram and the modified Dependency Matrix model which only uses one embedding for each word. To allow a direct comparison, we trained them all on the WikiWoods \cite{wikiwoods} snapshot of English Wikipedia. The same hyper-parameters were used: a dimensionality of 100,  15 negative samples, a batch size of 5, and an Adagrad optimiser with an initial learning rate of 0.025. For the dependency models, we used the universal dependency parser provided by \texttt{spaCy} \citep{spacy2}. We applied the two few-shot methods we devised, as well as the Additive model with window weighting, subsampling and negative sampling described in~\S\ref{sec:few-shot}. The hyperparameters were $t = 10^{-6}$, $k = 15$ and $n = 5$.

\subsection{Few-shot learning tasks}

\paragraph{Definitional Nonce (DN)} This task \citep{herbelot-baroni-2017-high} provides a single definition sentence for each test word. The test words are frequent words, which have gold vectors of high quality in the background space. At evaluation time, a new vector is computed for each test word, based on the few-shot learning model. The rank of the gold vector relatively to the inferred vector is then calculated, i.e.the number of words from the vocabulary which are closer to the inferred vector than the gold vector is. The distance metric is cosine similarity - the bigger the similarity, the smaller the distance. The metrics retrieved are the Mean Reciprocal Rank (MRR) and median rank. 

\paragraph{Chimera} The Chimera task \citep{lazaridou} provides non-existing words (chimeras) with 6 context sentences, as well as similarity scores between
the chimera and other existing words. The way in which the dataset was built simulates
few-shot learning for humans, since the participants of the experiment needed to infer the
meaning of a word they never saw before and rate its similarity with other concepts, based
only on the 6 context sentences.
Trials with 2, 4 and 6 context sentences are conducted. After each trial, the similarity scores
between the inferred vector and the vectors of the words provided is compared against the
human similarity scores by retrieving the Spearman’s rank correlation coefficient.

\paragraph{Contextual Rare Words (CRW)} Like Chimera, the CRW task \citep{khodak-etal-2018-la} is based
on human ratings between pairs of words. This time the pairs contain a rare word and a
frequent one, with an assumed reliable embedding in the background model. For each rare word, 255 context sentences are provided. The vector is generated using
the few-shot model for 1, 2, 4, 8, 16 context sentences, selected at random. For each such experiment, the similarity scores between the few-shot vector and
the background embedding of the non-rare word are calculated and compared against the human scores using the Spearman’s rank correlation coefficient. The scores are averaged out across 10 random selections of context sentences. 

\subsection{Results and Discussion}

The results in Table \ref{tab:results} show that the dependency-based background models performed better than window-based Skip-Gram on all three evaluation tasks. For all background models, applying the Dependency Additive technique consistently improved the results of the Additive model. For the DN task and DM background model, there were three cases where the  Additive model gave a rank of over 30,000, while the Dependency Additive model gave a rank of 1 or 2, showing the method's potential for sentences of specific structures. The DM additive model showed a promising result for the Chimera task, but was still outperformed by the Dependency Additive model, and its scores had the biggest variance across all combinations. This suggests that more careful weighting might be required.

\section{Conclusion}
\label{sec:conc}

We investigated the use of dependency information
for few-shot learning in distributional semantics.
We found that dependency-based contexts
are more useful than window-based contexts,
with better performance across three evaluation datasets.
We proposed a simplified version of the Dependency Matrix model, using only one vector per word,
which makes it easier to apply in a few-shot setting.

An important next step would be to investigate the use of the proposed methods for other languages, since our work was limited to English data and it is possible that the dependency structure is more relevant for few-shot learning in the case of specific languages. In order to do such an analysis, one would additionally need to create test data for the few shot-learning tasks, which would require the participation of speakers of the selected languages.

In future work, performance might be further improved
by training an A La Carte model (discussed in~\S\ref{sec:few-shot}),
where the use of dependencies
would make it possible to use a graph-convolutional network
\citep{marcheggiani-titov-2017-encoding}.

\bibliography{anthology,custom}

\begin{thebibliography}{21}
\expandafter\ifx\csname natexlab\endcsname\relax\def\natexlab#1{#1}\fi

\bibitem[{Adams et~al.(2017)Adams, Makarucha, Neubig, Bird, and
  Cohn}]{adams-etal-2017-cross}
Oliver Adams, Adam Makarucha, Graham Neubig, Steven Bird, and Trevor Cohn.
  2017.
\newblock \href {https://aclanthology.org/E17-1088} {Cross-lingual word
  embeddings for low-resource language modeling}.
\newblock In \emph{Proceedings of the 15th Conference of the {E}uropean Chapter
  of the Association for Computational Linguistics: Volume 1, Long Papers},
  pages 937--947, Valencia, Spain. Association for Computational Linguistics.

\bibitem[{Bertolini et~al.(2021)Bertolini, Weeds, Weir, and
  Peng}]{bertolini-etal-2021-representing}
Lorenzo Bertolini, Julie Weeds, David Weir, and Qiwei Peng. 2021.
\newblock \href {https://doi.org/10.18653/v1/2021.findings-acl.296}
  {Representing syntax and composition with geometric transformations}.
\newblock In \emph{Findings of the Association for Computational Linguistics:
  ACL-IJCNLP 2021}, pages 3343--3353, Online. Association for Computational
  Linguistics.

\bibitem[{Bojanowski et~al.(2017)Bojanowski, Grave, Joulin, and
  Mikolov}]{fasttext}
Piotr Bojanowski, Edouard Grave, Armand Joulin, and Tomas Mikolov. 2017.
\newblock \href {https://doi.org/10.1162/tacl_a_00051} {{Enriching Word Vectors
  with Subword Information}}.
\newblock \emph{Transactions of the Association for Computational Linguistics},
  5:135--146.

\bibitem[{Boleda(2020)}]{boleda2020distributional}
Gemma Boleda. 2020.
\newblock \href
  {https://www.annualreviews.org/doi/full/10.1146/annurev-linguistics-011619-030303}
  {Distributional semantics and linguistic theory}.
\newblock \emph{Annual Review of Linguistics}, 6:213--234.

\bibitem[{Czarnowska et~al.(2019)Czarnowska, Emerson, and
  Copestake}]{czarnowska-etal-2019-words}
Paula Czarnowska, Guy Emerson, and Ann Copestake. 2019.
\newblock \href {https://doi.org/10.18653/v1/W19-0408} {Words are vectors,
  dependencies are matrices: Learning word embeddings from dependency graphs}.
\newblock In \emph{Proceedings of the 13th International Conference on
  Computational Semantics - Long Papers}, pages 91--102, Gothenburg, Sweden.
  Association for Computational Linguistics.

\bibitem[{Emerson(2020)}]{emerson-2020-goals}
Guy Emerson. 2020.
\newblock \href {https://doi.org/10.18653/v1/2020.acl-main.663} {What are the
  goals of distributional semantics?}
\newblock In \emph{Proceedings of the 58th Annual Meeting of the Association
  for Computational Linguistics}, pages 7436--7453, Online. Association for
  Computational Linguistics.

\bibitem[{Flickinger et~al.(2010)Flickinger, Oepen, and
  Ytrest{\o}l}]{wikiwoods}
Dan Flickinger, Stephan Oepen, and Gisle Ytrest{\o}l. 2010.
\newblock \href
  {http://www.lrec-conf.org/proceedings/lrec2010/pdf/432_Paper.pdf}
  {Wiki{W}oods: Syntacto-semantic annotation for {E}nglish {W}ikipedia}.
\newblock In \emph{Proceedings of the 7th International Conference on Language
  Resources and Evaluation (LREC)}, pages 1665--1671. European Language
  Resources Association (ELRA).

\bibitem[{Herbelot and Baroni(2017)}]{herbelot-baroni-2017-high}
Aur{\'e}lie Herbelot and Marco Baroni. 2017.
\newblock \href {https://doi.org/10.18653/v1/D17-1030} {High-risk learning:
  acquiring new word vectors from tiny data}.
\newblock In \emph{Proceedings of the 2017 Conference on Empirical Methods in
  Natural Language Processing}, pages 304--309, Copenhagen, Denmark.
  Association for Computational Linguistics.

\bibitem[{Honnibal et~al.(2020)Honnibal, Montani, Van~Landeghem, and
  Boyd}]{spacy2}
Matthew Honnibal, Ines Montani, Sofie Van~Landeghem, and Adriane Boyd. 2020.
\newblock \href {https://doi.org/10.5281/zenodo.1212303} {{spaCy}:
  Industrial-strength natural language processing in {P}ython}.

\bibitem[{Khodak et~al.(2018)Khodak, Saunshi, Liang, Ma, Stewart, and
  Arora}]{khodak-etal-2018-la}
Mikhail Khodak, Nikunj Saunshi, Yingyu Liang, Tengyu Ma, Brandon Stewart, and
  Sanjeev Arora. 2018.
\newblock \href {https://doi.org/10.18653/v1/P18-1002} {A la carte embedding:
  Cheap but effective induction of semantic feature vectors}.
\newblock In \emph{Proceedings of the 56th Annual Meeting of the Association
  for Computational Linguistics (Volume 1: Long Papers)}, pages 12--22,
  Melbourne, Australia. Association for Computational Linguistics.

\bibitem[{Lazaridou et~al.(2017)Lazaridou, Marelli, and Baroni}]{lazaridou}
Angeliki Lazaridou, Marco Marelli, and Marco Baroni. 2017.
\newblock \href {https://doi.org/10.1111/cogs.12481} {Multimodal word meaning
  induction from minimal exposure to natural text}.
\newblock \emph{Cognitive Science}, 41 Suppl 4.

\bibitem[{Lenci(2018)}]{lenci2018distributional}
Alessandro Lenci. 2018.
\newblock \href
  {https://www.annualreviews.org/doi/full/10.1146/annurev-linguistics-030514-125254}
  {Distributional models of word meaning}.
\newblock \emph{Annual review of Linguistics}, 4:151--171.

\bibitem[{Levy and Goldberg(2014)}]{levy-goldberg-2014-dependency}
Omer Levy and Yoav Goldberg. 2014.
\newblock \href {https://doi.org/10.3115/v1/P14-2050} {Dependency-based word
  embeddings}.
\newblock In \emph{Proceedings of the 52nd Annual Meeting of the Association
  for Computational Linguistics (Volume 2: Short Papers)}, pages 302--308,
  Baltimore, Maryland. Association for Computational Linguistics.

\bibitem[{Levy et~al.(2015)Levy, Goldberg, and
  Dagan}]{levy-etal-2015-improving}
Omer Levy, Yoav Goldberg, and Ido Dagan. 2015.
\newblock \href {https://doi.org/10.1162/tacl_a_00134} {Improving
  distributional similarity with lessons learned from word embeddings}.
\newblock \emph{Transactions of the Association for Computational Linguistics},
  3:211--225.

\bibitem[{Marcheggiani and Titov(2017)}]{marcheggiani-titov-2017-encoding}
Diego Marcheggiani and Ivan Titov. 2017.
\newblock \href {https://doi.org/10.18653/v1/D17-1159} {Encoding sentences with
  graph convolutional networks for semantic role labeling}.
\newblock In \emph{Proceedings of the 2017 Conference on Empirical Methods in
  Natural Language Processing}, pages 1506--1515, Copenhagen, Denmark.
  Association for Computational Linguistics.

\bibitem[{Mikolov et~al.(2013)Mikolov, Sutskever, Chen, Corrado, and
  Dean}]{mikolov}
Tomas Mikolov, Ilya Sutskever, Kai Chen, Greg~S Corrado, and Jeff Dean. 2013.
\newblock \href
  {http://papers.nips.cc/paper/5021-distributed-representations-of-words-and-phrases-and-their-compositionality.pdf}
  {Distributed representations of words and phrases and their
  compositionality}.
\newblock In C.~J.~C. Burges, L.~Bottou, M.~Welling, Z.~Ghahramani, and K.~Q.
  Weinberger, editors, \emph{Advances in Neural Information Processing Systems
  26}, pages 3111--3119. Curran Associates, Inc.

\bibitem[{Perez et~al.(2021)Perez, Kiela, and Cho}]{perez2021true}
Ethan Perez, Douwe Kiela, and Kyunghyun Cho. 2021.
\newblock \href
  {https://proceedings.neurips.cc/paper/2021/hash/5c04925674920eb58467fb52ce4ef728-Abstract.html}
  {True few-shot learning with language models}.
\newblock In \emph{Advances in Neural Information Processing Systems}.

\bibitem[{Sahlgren(2008)}]{sahlgren2008distributional}
Magnus Sahlgren. 2008.
\newblock \href
  {https://www.diva-portal.org/smash/get/diva2:1041938/FULLTEXT01.pdf} {The
  distributional hypothesis}.
\newblock \emph{Italian Journal of Disability Studies}, 20:33--53.

\bibitem[{Schick and Sch{\"u}tze(2019)}]{schick2019learning}
Timo Schick and Hinrich Sch{\"u}tze. 2019.
\newblock \href {https://ojs.aaai.org/index.php/AAAI/article/view/4675/4553}
  {Learning semantic representations for novel words: Leveraging both form and
  context}.
\newblock In \emph{Proceedings of the AAAI Conference on Artificial
  Intelligence}, volume~33, pages 6965--6973.

\bibitem[{Van~Hautte et~al.(2019)Van~Hautte, Emerson, and
  Rei}]{van-hautte-etal-2019-bad}
Jeroen Van~Hautte, Guy Emerson, and Marek Rei. 2019.
\newblock \href {https://doi.org/10.18653/v1/D19-6104} {Bad form: Comparing
  context-based and form-based few-shot learning in distributional semantic
  models}.
\newblock In \emph{Proceedings of the 2nd Workshop on Deep Learning Approaches
  for Low-Resource NLP (DeepLo 2019)}, pages 31--39, Hong Kong, China.
  Association for Computational Linguistics.

\bibitem[{Zobnin and Elistratova(2019)}]{zobnin-elistratova-2019-learning}
Alexey Zobnin and Evgenia Elistratova. 2019.
\newblock \href {https://doi.org/10.18653/v1/W19-4329} {Learning word
  embeddings without context vectors}.
\newblock In \emph{Proceedings of the 4th Workshop on Representation Learning
  for NLP (RepL4NLP-2019)}, pages 244--249, Florence, Italy. Association for
  Computational Linguistics.

\end{thebibliography}
\bibliographystyle{acl_natbib}

\appendix
\end{document}